\documentclass[conference]{IEEEtran}
\IEEEoverridecommandlockouts
\usepackage{multirow}
\usepackage{cite}
\usepackage{amsmath,amssymb,amsfonts}
\usepackage{algorithmic}
\usepackage{graphicx}
\usepackage{array}
\usepackage{textcomp}
\usepackage{booktabs}
\usepackage{xcolor,colortbl}
\usepackage{array}
\usepackage{cite}
\usepackage{pifont}
\usepackage{amsmath,amssymb,amsfonts}
\usepackage{algorithmic}
\usepackage{graphicx}
\usepackage{textcomp}
\usepackage{xcolor}
\usepackage{makecell}

\def\BibTeX{{\rm B\kern-.05em{\sc i\kern-.025em b}\kern-.08em
    T\kern-.1667em\lower.7ex\hbox{E}\kern-.125emX}}
\begin{document}

\title{{MedCoAct: Confidence-Aware Multi-Agent Collaboration for Complete Clinical Decision}}
\author{
\IEEEauthorblockN{Hongjie Zheng\IEEEauthorrefmark{1}, 
Zesheng Shi\IEEEauthorrefmark{2}, 
Ping Yi\IEEEauthorrefmark{1}\thanks{Corresponding author}}
\IEEEauthorblockA{\IEEEauthorrefmark{1}\textit{Shanghai Jiao Tong University}, 
\IEEEauthorrefmark{2}\textit{Harbin Institute of Technology}}
}
\maketitle

\begin{abstract}

Autonomous agents utilizing Large Language Models (LLMs) have demonstrated remarkable capabilities in isolated medical tasks like diagnosis and image analysis, but struggle with integrated clinical workflows that connect diagnostic reasoning and medication decisions. 
We identify a core limitation: existing medical AI systems process tasks in isolation without the cross-validation and knowledge integration found in clinical teams, reducing their effectiveness in real-world healthcare scenarios.
To transform the isolation paradigm into a collaborative approach, we propose MedCoAct, a confidence-aware multi-agent framework that simulates clinical collaboration by integrating specialized doctor and pharmacist agents, and present a benchmark, DrugCareQA, to evaluate medical AI capabilities in integrated diagnosis and treatment workflows.
Our results demonstrate that MedCoAct achieves 67.58\% diagnostic accuracy and 67.58\% medication recommendation accuracy, outperforming single agent framework by 7.04\% and 7.08\% respectively. 
This collaborative approach generalizes well across diverse medical domains, proving especially effective for telemedicine consultations and routine clinical scenarios, while providing interpretable decision-making pathways.
\end{abstract}

\vspace{-5mm}
\section{Introduction}
Autonomous agents utilizing Large Language Models (LLMs) show promise in enhancing complex professional tasks. In healthcare, LLMs have demonstrated capabilities across diverse applications. 
Medical question-answering systems provide rapid access to comprehensive clinical knowledge and evidence-based recommendations \cite{liu2025distillationpushinglimitsmedical, yue2024clinicalagentclinicaltrialmultiagent, nori2025sequentialdiagnosislanguagemodels}. 
LLMs assist also with medical imaging report generation, significantly reducing physician workload \cite{DBLP:conf/aaai/LiuTCS024}.
Moreover, LLMs help drug discovery research by accelerating molecular design and optimization processes \cite{DBLP:journals/corr/abs-2012-09355}. 

However, existing systems struggle with complex medical workflows requiring integrated diagnosis and medication decisions. 
Although multi-agent systems have been explored for medical AI, existing frameworks mostly focus on single tasks rather than complex integrated workflows \cite{tang2024medagentslargelanguagemodels}.
Moreover, they lack reflective mechanisms for dynamic quality optimization, leading to diagnostic errors and suboptimal medication recommendations, with hallucinated information propagating through subsequent reasoning steps. 
\begin{figure}[t]
    \centering
    \includegraphics[width=0.5\textwidth]{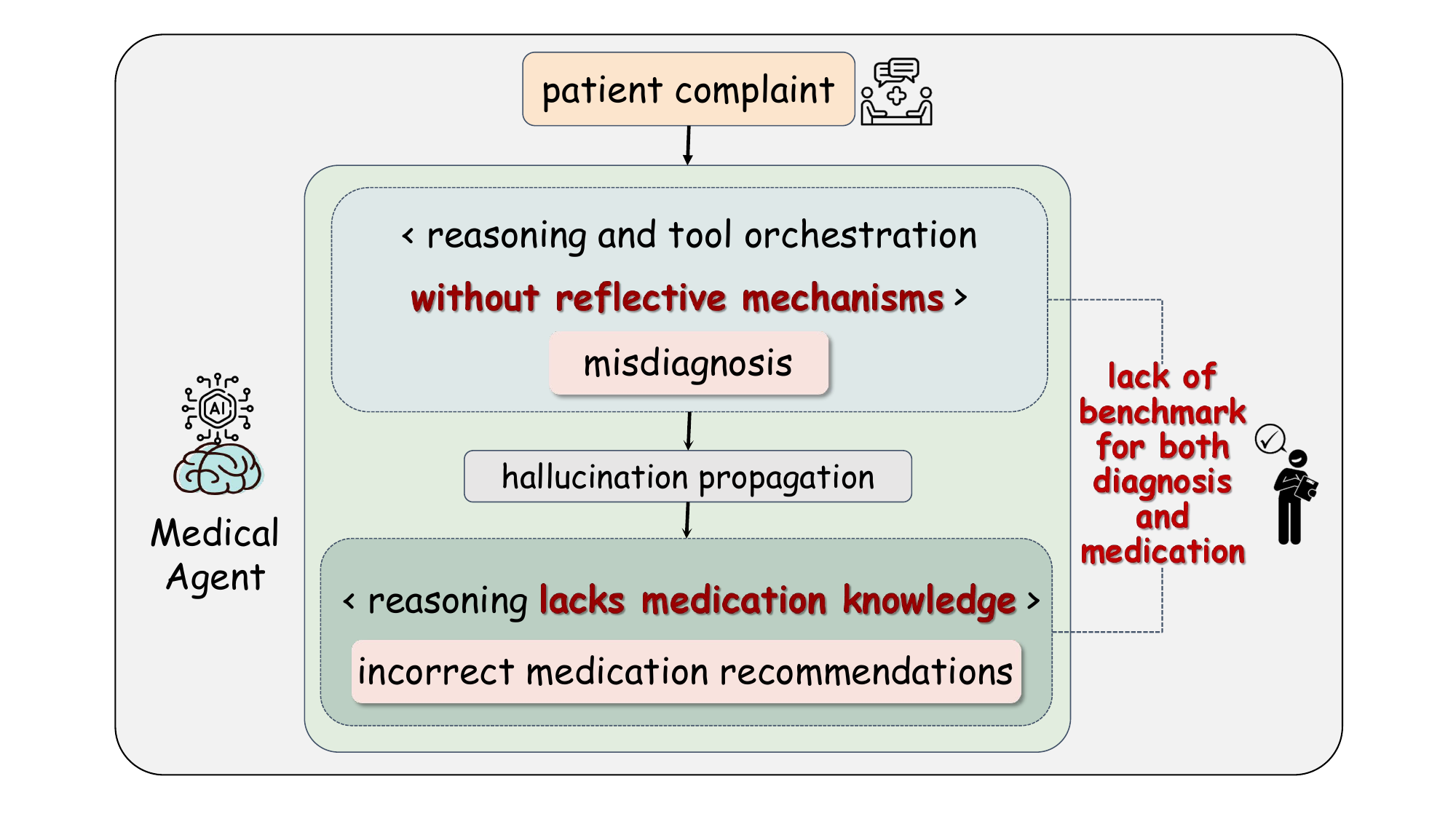}
    \caption{Current medical agents lack reflective mechanisms, causing misdiagnosis, hallucination propagation, and incorrect medication recommendations. It exists also the absence of benchmarks for joint diagnosis-medication evaluation.}
    \vspace{-7mm}
    \label{fig:problem}
\end{figure}
To advance medical AI systems, numerous benchmark construction methods have been proposed \cite{DBLP:conf/nips/LiBFIPKT24}, yet current benchmark datasets exhibit similar limitations, typically covering isolated tasks rather than complete medical decision-making processes.

To address these gaps, we introduce the \textbf{DrugCareQA} dataset and propose \textbf{MedCoAct} (Medical Collaborative Action), a dual agent system for integrated medical decision making. 
DrugCareQA contains 2,700 medical consultation cases covering integrated diagnosis and medication decision making. 
MedCoAct significantly benefits from doctor and pharmacist role specialization which aligns with medical practice. 
Each role possesses deeper expertise within their specialized domains, thereby improving overall accuracy.
Additionally, the system incorporates confidence-aware reflection mechanisms that enable self-assessment of tool invocation results and decision certainty. Furthermore, adaptive retrieval strategies dynamically adjust knowledge sourcing based on specific scenarios, improving retrieval quality.

Experimental results show MedCoAct achieves 67.58\% Top-1 diagnostic accuracy and 67.58\% medication recommendation accuracy on DrugCareQA, outperforming single agent system by 7.04\% and 7.08\% respectively. These improvements validate the effectiveness of specialized agent collaboration in medical decision making.

Our main contributions can be summarized as follows:
\begin{itemize}
\item \textbf{DrugCareQA comprehensive benchmark}: A dataset with 2,700 real-world medical consultation cases covering diagnosis-to-medication workflows, enabling comprehensive evaluation of integrated medical AI systems.
\item \textbf{Integrated diagnosis-medication framework}: MedCoAct addresses medical workflows through specialized doctor and pharmacist agent collaboration, bridging diagnosis and medication decision making in unified system.
\item \textbf{Confidence aware reflection mechanism}: Adaptive reflection enabling agents to autonomously assess and optimize decisions when confidence is low, enhancing decision accuracy and safety.
\item \textbf{Specialized retrieval strategies}: Role customized knowledge acquisition providing targeted professional support for agents through adaptive retrieval mechanisms.
\end{itemize}

\vspace{-1mm}
\section{Related work}
\subsection{LLM for medicine}\label{AA}
LLMs address complex reasoning and decision support challenges in medical diagnosis \cite{pandey2024advancinghealthcareautomationmultiagent, DBLP:conf/nips/WangCQCCJZNN24}. 
Prompt engineering methods enable LLMs to simulate physician reasoning processes, demonstrating strong performance in reasoning transparency for complex cases\cite{DBLP:conf/aaai/KwonOKMLHSSLY24}. 
Wu et al. demonstrated LLMs' potential in clinical diagnostic reasoning. Kwon et al. proposed the Clinical Chain-of-Thought (CoT) framework for generating interpretable diagnostic reasoning pathways.

However, inherent knowledge limitations of LLMs may cause incompleteness or inaccuracy issues. Retrieval-Augmented Generation (RAG) techniques have been applied to address this challenge\cite{jiang2024graphcareenhancinghealthcarepredictions}. MedGraphRAG proposed by Wu et al. using graph algorithms for medical knowledge integration\cite{wu2024medicalgraphragsafe}. MedRAG developed by Zhao et al. employs diagnostic knowledge graphs with Electronic Health Record (EHR) data, enhancing medical reasoning capabilities\cite{DBLP:conf/www/ZhaoLYM25}. Lu et al. propose ClinicalRAG utilizing heterogeneous knowledge retrieval strategies to reduce medical misinformation propagation\cite{lu-etal-2024-clinicalrag}.

To enhance conversational interaction and reasoning abilities, fine-tuning and specialized pre-training have emerged as another technical pathway\cite{DBLP:journals/eaai/KraisnikovicHPZHM25}. The Zhongjing model proposed by Yang et al. employs comprehensive training combining continual pre-training, Supervised Fine-Tuning (SFT), and Reinforcement Learning from Human Feedback (RLHF) with the CMtMedQA dataset, improving multi-turn diagnostic dialogue and proactive inquiry capabilities\cite{yang2023zhongjingenhancingchinesemedical}. The Citrus model developed by Wang et al. demonstrates superior reasoning performance in complex differential diagnosis tasks \cite{wang2025citrusleveragingexpertcognitive}.
\begin{figure*}[t]
    \centering
    \includegraphics[width=\textwidth]{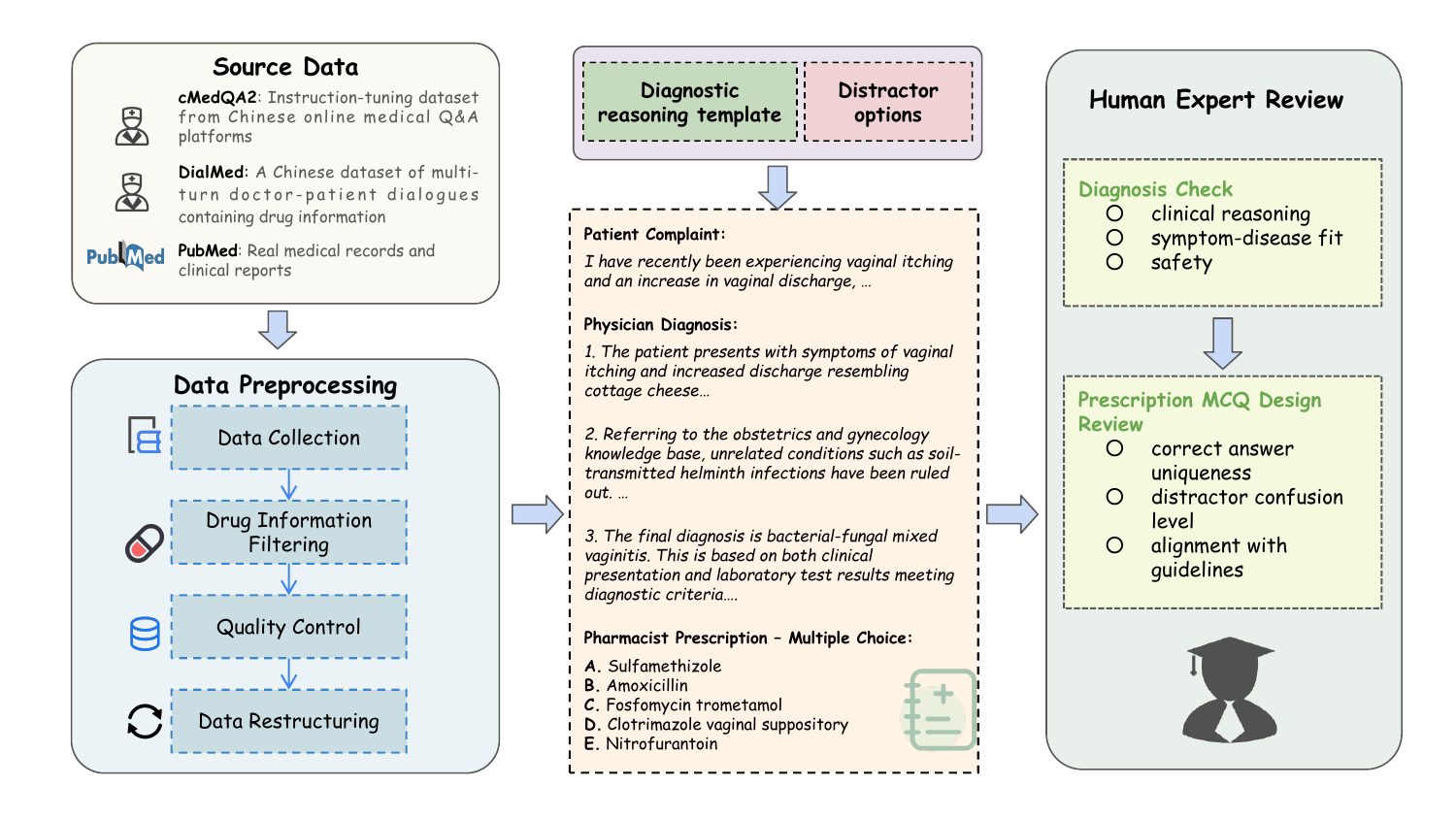}
    \vspace{-12mm}
    \caption{Overview of the DrugCareQA benchmark construction pipeline. The workflow consists of data collection, quality control, annotation process, and evaluation metric design.}
    \label{fig:drugcareqa-workflow}
\end{figure*}
\vspace{-6mm}
\subsection{LLM-based multi-agent frameworks}
LLM-based multi-agent systems have gained attention and demonstrated advantages in professional domains\cite{DBLP:journals/corr/abs-2306-03314}. Tang et al. proposed MedAgents which improves medical knowledge response accuracy through iterative discussions among domain expert agents until consensus is reached. 
Zhou Yuan et al. introduced the ZODIAC multi-agent framework, which enhances cardiovascular diagnostic professionalism and fairness through collaboration among multimodal LLM agents and expert validation mechanisms\cite{zhou2024zodiaccardiologistlevelllmframework}. However, multi-agent systems still exhibit performance limitations in complex scenarios. 
To develop optimized workflows, Hong et al.
integrated Standardized Operating Procedures (SOPs) into prompt sequences, achieving streamlined workflows and improved performance by simulating real human collaboration patterns\cite{DBLP:conf/iclr/HongZCZCWZWYLZR24}. 
Zhang et al., through improved Monte Carlo Tree Search (MCTS) and dual optimization that mimics human cognitive flexibility for dynamic adaptation in complex web environments\cite{zhang2024webpilotversatileautonomousmultiagent}.
To leverage multi-agent framework advantages, Thang Nguyen et al. proposed the MA-RAG framework\cite{nguyen2025maragmultiagentretrievalaugmentedgeneration}. This framework improves reasoning precision and retrieval accuracy without fine-tuning through collaborative CoT reasoning mechanisms.
\vspace{-5mm}
\section{Benchmark}
\vspace{-1mm}
With the widespread adoption of online medical consultation platforms, developing AI systems for real patient-doctor dialogue scenarios has become increasingly important. 
However, existing medical AI benchmarks exhibit significant limitations in effectively evaluating model performance in actual consultation scenarios. 

As shown in Table~\ref{tab:medical_benchmarks}, mainstream benchmarks such as MedQA\cite{DBLP:journals/corr/abs-2410-01553}, PubMedQA\cite{DBLP:conf/emnlp/JinDLCL19}, and MedMCQA\cite{DBLP:conf/chil/PalUS22} are primarily based on medical examination or literature abstracts, lacking authentic patient-doctor interaction data. 
Moreover, these benchmarks typically employ multiple-choice question formats that fail to capture the ambiguity and diversity of symptom descriptions in real consultations\cite{DBLP:journals/corr/abs-2412-15194}. 
Additionally, these benchmarks focus solely on diagnostic capability assessment while overlooking medication recommendation which is a critical component of clinical decision \cite{chen2025benchmarkinglargelanguagemodels}.

Besides, although the recently introduced HealthBench incorporates multi-turn dialogue formats, it targets high-complexity scenarios beyond typical consultations, and its data sourced from hospitals differs significantly from natural patient expressions \cite{DBLP:journals/corr/abs-2505-08775}. Furthermore, existing benchmarks don't integrate diagnosis and medication into unified evaluation, which makes comprehensive assessment of medical consultation capabilities challenging.

To address these challenges, we develop DrugCareQA, a comprehensive benchmark dataset specifically designed for real online medical consultation scenarios. 
Compared to existing benchmarks, DrugCareQA offers three key innovations: 
\begin{itemize}
\item We represent a medical benchmark constructed from authentic dialogues between patients and doctors.
\item We integrate diagnostic reasoning with medication selection into a unified evaluation framework.
\item The dataset encompasses 2,700 annotated cases across seven clinical departments with dual quality verification combining knowledge base validation and expert review.
\end{itemize}

\vspace{-3mm}
\begin{table*}[htbp]
\centering
\caption{Medical AI Benchmarks Dataset Overview}
\vspace{-1mm}
\label{tab:medical_benchmarks}
\resizebox{\linewidth}{!}{
\begin{tabular}{lcccc}
\toprule
\textbf{Benchmark} & \textbf{Answer Format} & \textbf{Domain} & \textbf{Sample Size} & \textbf{Source Type} \\
\midrule
MedQA & 4-option MCQs & Medical knowledge understanding and reasoning & 1,273 & Examination \\
\addlinespace[0.1cm]
PubMedQA & 3-option MCQs & Biomedical research text understanding and reasoning & 1,000 & Literature \\
\addlinespace[0.1cm]
MedMCQA & 4-option MCQs & Comprehensive medical knowledge Q\&A & 4,183 & Examination \\
\addlinespace[0.1cm]
MedBullets & 5-option MCQs & Clinical medicine Q\&A & 308 & Examination \\
\addlinespace[0.1cm]
MMLU\cite{DBLP:conf/iclr/HendrycksBBZMSS21} & 4-option MCQs & Large-scale multitask medical knowledge assessment & 1,871 & Examination \\
\addlinespace[0.1cm]
MMLU-Pro\cite{DBLP:conf/nips/WangMZNCGRAHJLK24} & 10-option MCQs & High-difficulty multitask medical knowledge challenge & 818 & Examination \\
\addlinespace[0.1cm]
CareQA\cite{DBLP:conf/bibm/XiaoWTLS24} & 4-option MCQs & Professional medical training knowledge Q\&A & 5,410 & Examination \\
\addlinespace[0.1cm]
JMed & 21-option MCQs & Real clinical data simulation diagnosis & 1,000 & Clinical cases \\
\addlinespace[0.1cm]
HealthBench & Multi-turn dialogue & Comprehensive medical scenarios and professional consultation & 5,000 & Hospital data \\
\addlinespace[0.1cm]
\midrule
DrugCareQA (ours) & Diagnostic Q\&A and drug selection & Real medical record clinical diagnosis and drug decision-making & 2,700 & Clinical cases \\
\bottomrule
\end{tabular}}
\vspace{-5mm}
\end{table*}

\subsection{Multi-source medical data collection}

We adopt a dual source strategy combining real clinical dialogues from chinese online medical platforms with authoritative PubMed literature. 
Online platform data consists of screened patient and physician conversations of moderate length containing complete diagnostic information \cite{DBLP:journals/access/ZhangZWGL18,DBLP:conf/coling/HeHOGCX022}. 
PubMed literature includes peer reviewed clinical reports and case studies with explicit diagnostic content and clinical guidelines. 
These sources offer complementary strengths: PubMed provides
standardized clinical cases with typical presentations, while online dialogues reflect authentic clinical communication.
This integrated approach ensures both clinical authenticity and scientific validity, providing multidimensional data for medical AI evaluation.
\vspace{-2mm}
\subsection{Drug-centered data screening and standardization}
\vspace{-1mm}
To evaluate the complete medical workflow from patient complaints to diagnosis and prescription, we screen high-quality records containing all three components.
For online medical platform data, we employ LLMs to automatically extract patient-physician dialogues containing complete prescription information. 
We transform the unstructured dialogue data into a standardized three-column format: chief complaint, diagnosis, and medication. 
For PubMed literature data, we systematically extract key clinical elements including patient demographics, clinical presentations, diagnostic processes, final diagnoses, and therapeutic regimens.
Through automated text processing methods, we standardize all extracted information into the same three-column framework, providing a consistent foundation for computational analysis.

\subsection{Medical knowledge base-driven data quality control}

To ensure medical accuracy, we built a validation benchmark from 300 authoritative medical textbooks.
We designed a multi-layer verification mechanism: first retrieve relevant medical documents based on semantic similarity with target data, then employ LLMs for medical fact consistency comparisons. 
This mechanism automatically identifies medical errors, filtering reliable data through confidence scoring.
\vspace{-3mm}
\subsection{Clinical thinking-guided question construction}
\vspace{-1mm}
To construct data reflecting clinical thinking processes, we deconstructed the diagnostic reasoning process into a three-stage cognitive framework of symptom analysis, step-by-step reasoning, and final decision-making. 
Under the guidance of medical experts, we meticulously designed multi-level prompt engineering strategies that enable LLMs to simulate authentic physician diagnostic thinking \cite{wu2024guidingclinicalreasoninglarge}. 
In option design, we skillfully incorporated common clinical practice pitfalls and similar drug distractors, constructing hierarchical difficulty gradients. 
Each question not only evaluates disease and pharmaceutical knowledge memorization but also assesses clinical logic.
\vspace{-1mm}
\subsection{Expert review and quality assurance system}  
\vspace{-1mm}
Considering strict medical accuracy needs, we establish a multi-level expert review mechanism. 
Initially, three medical experts independently evaluated the medical accuracy and medication appropriateness of each question, focusing on the rigor of diagnostic logic and the suitability of drug selection. 
Subsequently, one senior medical expert was invited to provide final adjudication on disputed questions, ensuring that the diagnostic reasoning process adheres to clinical practice standards and that medication recommendations match patient conditions. 
Through this quality control process, we ensured the professionalism and reliability of the dataset.
\vspace{-1mm}
\section{MedCoAct framework}
\vspace{-1mm}
\begin{figure*}[t]
    \centering
    \includegraphics[width=\textwidth]{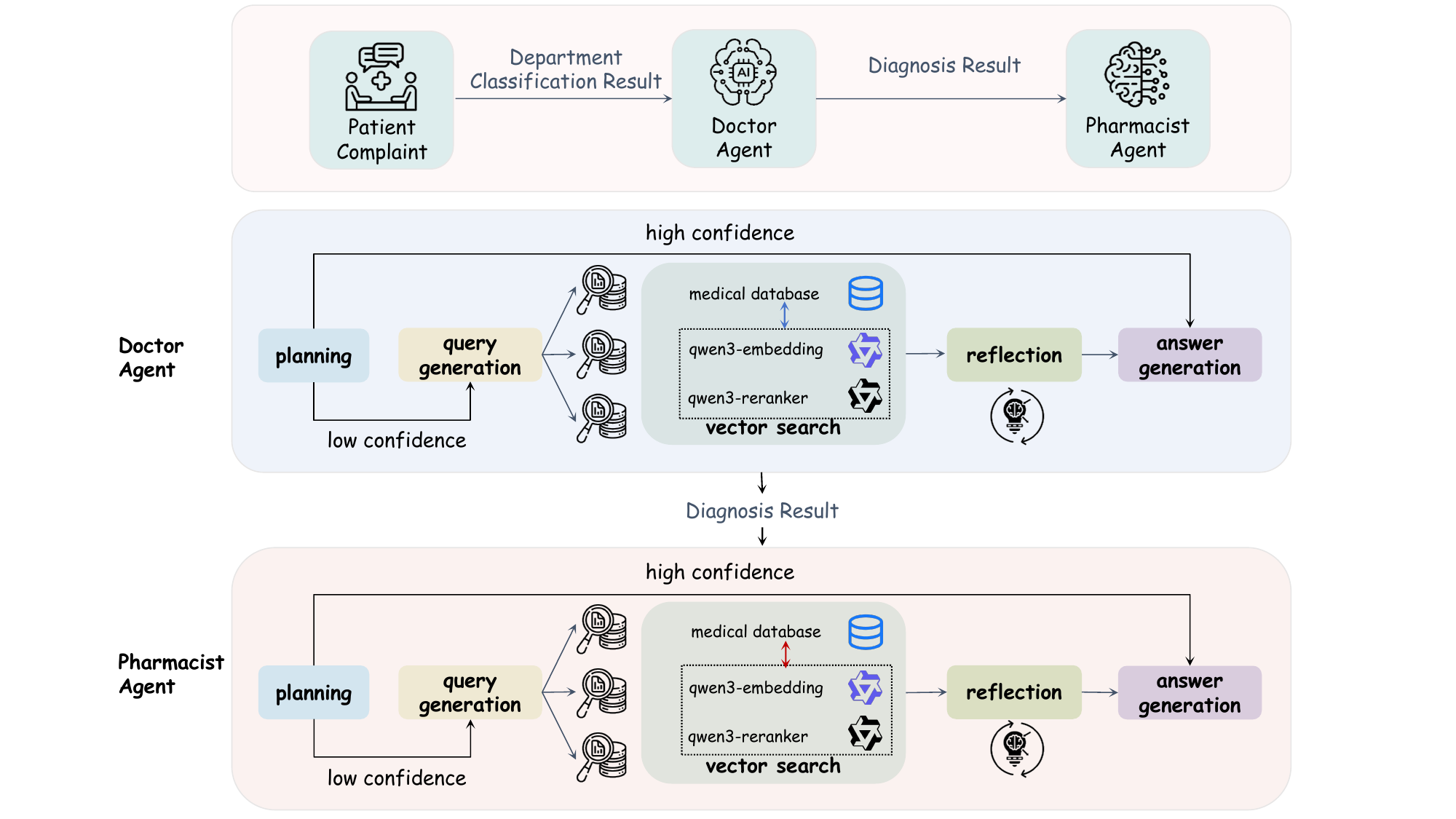}
    \caption{The framework demonstrates a complete workflow from patient complaints through doctor agent diagnosis to pharmacist agent medication recommendations. 
    Both agents employ the same five-step architecture of planning, query generation, knowledge retrieval, reflection, and answer generation.
    The system incorporates confidence mechanisms, multi-path intelligent query retrieval, vector search tools, and reflection mechanisms to enable cross-agent collaboration and improve medical accuracy.}
    \vspace{-6mm}
    \label{fig:MedCoAct framework}
\end{figure*}
However, agents may have significant limitations during processing: they sometimes generate inappropriate queries, making it difficult to locate accurate reference documents. 
Meanwhile, lacking effective confidence estimation of outputs generated by other agents leads to hallucination propagation and inappropriate medication recommendations.

To address these challenges, the MedCoAct framework develops a collaborative mechanism, enabling cooperation between doctor agent and pharmacist agent for diagnosis and prescription. 
In this section, firstly we elaborate on role specialization and cross-agent workflow, showing how specialized roles and collaborative mechanisms enhance diagnostic and prescription accuracy. 
Second, we present the query planning and reflection mechanism, enabling adaptive query generation and automatic re-optimization when confidence is low \cite{DBLP:journals/corr/abs-2405-06682}. 
Finally, we introduce a vector search tool framework addressing inaccurate and incomplete document retrieval.
\vspace{-2.5mm}
\subsection{Role specialization and cross-agent workflow}
\vspace{-1mm}
\subsubsection{Role specialization}
Complex medical problems require collaboration among different expertise. 
Following real-world clinical practice where doctors handle diagnosis and pharmacists manage medication, we define two core roles: \textbf{doctor} agent and \textbf{pharmacist} agent.

Through carefully designed prompt engineering, we inject specific clinical thinking patterns and professional reasoning capabilities into each agent, enabling diagnostic reasoning approach for the \textbf{doctor} agent and prescribing reasoning approach for the \textbf{pharmacist} agent.

The \textbf{doctor} agent generates structured diagnostic planning from patient complaints and optimizes medical literature retrieval through adaptive strategies. 
When handling complaints, it executes department classification based on symptoms and medical histories, generates multiple targeted queries for medical literature, and dynamically modulates retrieval through document utility scoring and confidence monitoring. 
The \textbf{pharmacist} agent generates diversified therapeutic queries from diagnostic conclusions and patient symptoms, targeting pharmaceutical literature including drug indications, contraindications, interactions, and dosage recommendations.

\subsubsection{Cross-Agent Workflow}
We design a collaborative workflow that connects diagnosis and prescription. 
Specifically, after receiving patient queries, the \textbf{doctor} agent first conducts symptom analysis and department classification, formulating structured diagnostic planning. 
The diagnostic output is subsequently passed to the \textbf{pharmacist} agent, which receives the diagnostic results from the \textbf{doctor} agent and autonomously evaluates whether to adopt the diagnostic information for pharmaceutical treatment plan formulation. 
The \textbf{pharmacist} agent makes decisions based on its own professional judgment, combining patient symptoms and diagnostic conclusions to generate personalized medication recommendations. 
This collaborative mechanism ensures effective information transmission from diagnosis to medication while maintaining the independence and professionalism of the \textbf{pharmacist} agent in medication decision making, ultimately generating precise medical advice and medication regimens.
\vspace{-2mm}
\subsection{Query planning and reflection mechanism}
\vspace{-1mm}
In medical agent systems, precise and comprehensive query generation is crucial for acquiring accurate medical knowledge. 
Both \textbf{doctor} agents and \textbf{pharmacist} agents must leverage all available patient information to query relevant medical literature as comprehensively and accurately as possible, maintaining decision safety and reliability.

\subsubsection{Query planning}
Regarding query generation, we design specialized query planning strategies tailored to different agent roles. 
The \textbf{doctor} agent performs department classification based on patient symptoms and histories, then generates multiple targeted search queries to retrieve relevant medical guidelines. 
The system selects specialized prompt templates according to agent role, creating specific instructions to guide the vector search tool's retrieval process. 
Specifically, the prompt template for the \textbf{doctor} agent not only incorporates department classification guidance but also focuses on core medical concepts such as disease symptoms, diagnostic criteria, and differential diagnosis. 
The \textbf{pharmacist} agent generates targeted search queries for therapeutic regimens based on diagnostic results and patient symptoms. Its prompt template incorporates key pharmaceutical knowledge: drug mechanisms, indications, contraindications, and interactions. The agent automatically creates multiple queries to retrieve relevant pharmaceutical guideline documents.

\subsubsection{Reflection mechanism}
Agents may obtain insufficient or low-quality documents during initial retrieval, potentially compromising subsequent reasoning. 
To solve this challenge, we introduce a confidence-aware reflection mechanism enabling agents to evaluate retrieval quality and perform iterative optimization\cite{NEURIPS2024_fa54b0ed}.
When confidence falls below a threshold, agents automatically revert to planning and regenerate improved queries. This mechanism efficiently manages resources while enhancing diagnostic accuracy and medication safety.
\vspace{-1mm}
\subsection{Knowledge retrieval and vector search tool framework}
\vspace{-1mm}
To support precise knowledge acquisition for multi-agent collaborative diagnostic systems, we construct a specialized vector retrieval framework as the core knowledge acquisition component for agents.
\paragraph{Vector search tool architecture}
We propose a domain-driven database construction method that intelligently classifies documents based on professional domain characteristics. 
Documents solely focusing on disease symptoms, pathology, or diagnosis are assigned to the doctor knowledge base, while those exclusively focusing on medication guidelines, dosages, or precautions go to the pharmacist knowledge base. 
Most comprehensive documents containing both diagnostic and therapeutic information use dual indexing and are stored in both databases while preserving semantic integrity. 
The system then segments documents into text chunks and converts them into semantic vector representations through embedding model.
The vector search tool adopts a two-stage retrieval architecture for role-aware document retrieval\cite{monir2024vectorsearchenhancingdocumentretrieval}. 
In coarse retrieval, the system dynamically selects specialized templates based on the agent's role, generating domain-specific instructions and query vectors to guide Qwen3-embedding in recalling Top-K candidates from the corresponding knowledge base\cite{zhang2025qwen3embeddingadvancingtext}. 
In fine-grained reranking, the system generates customized instructions and queries for Qwen3-reranker to integrate role-specific preferences, clinical utility, and contextual relevance, outputting Top-N high-confidence documents.

\paragraph{Knowledge acquisition process}
Agents proactively invoke the vector retrieval tool based on professional requirements and task states to obtain relevant medical guideline documents. 
The system performs multi-dimensional confidence evaluation on retrieval results, including assessments of information sufficiency and accuracy. 
When confidence falls below the predetermined threshold, agents employ query reconstruction strategies, iteratively optimizing retrieval parameters and re-invoking the retrieval component to ensure acquisition of high-quality medical knowledge support that meets diagnostic reasoning requirements.
\vspace{-1mm}
\section{Results}
\vspace{-2mm}
\subsection{Experimental setting}
\paragraph{Dataset}
We evaluate the performance with the DrugCareQA dataset, covering two core medical tasks: diagnosis and medication selection.
\paragraph{Evaluation metrics}
We assess the task accuracy of clinical systems through the following core metrics: 
\begin{itemize}
\item\textbf{Top-1 diagnostic accuracy}: the proportion of cases where the system's primary diagnostic result matches the standard answer.
\item\textbf{Top-3 diagnostic accuracy}: the proportion of cases where the standard answer appears among the system's top three diagnostic suggestions.
\item\textbf{Drug prescription accuracy}: the accuracy of the system's recommended medication treatment regimens matching standard prescriptions. 
\end{itemize}

Additionally, to assess document retrieval quality, we employ two dimensions: \textbf{Relevance} and \textbf{Contribution}. 
\begin{itemize}
\item\textbf{Relevance}: Relevance measures how well documents semantically and topically align with patient questions, including symptoms, conditions, or scenarios. 
Higher relevance indicates more accurate addressing of patient core concerns, providing focused information on symptoms, diagnoses, and treatments.
\item\textbf{Contribution}: Contribution measures how well retrieved documents support agents in generating correct medical answers, evaluated against gold-standard responses. Higher contribution indicates more effective guidance for agent reasoning, providing key information for accurate diagnoses or treatment suggestions.
\end{itemize}
We adopt a maximum value approach to calculate these scores, as even a single quality document can provide crucial diagnostic insights in medical contexts.

\paragraph{Baseline methods}
We compare MedCoAct against two baseline methods: 
\begin{itemize}
\item\textbf{Simple agentic RAG system}: employs autonomous agents that can dynamically decide when to utilize vector search tool, completing diagnosis and prescription through intelligent query routing and adaptive retrieval.
\item\textbf{Local deep research system}: adopts an iterative web research workflow\cite{huang2025deepresearchagentssystematic}. It first converts patient complaints into optimized search queries, then conducts web search and organizes results. Based on retrieved information, it performs reflective analysis to identify knowledge gaps, generates follow-up queries, and ultimately produces diagnostic and treatment outcomes.
\end{itemize}
\begin{figure}[t]
    \centering
    \includegraphics[width=0.45\textwidth]{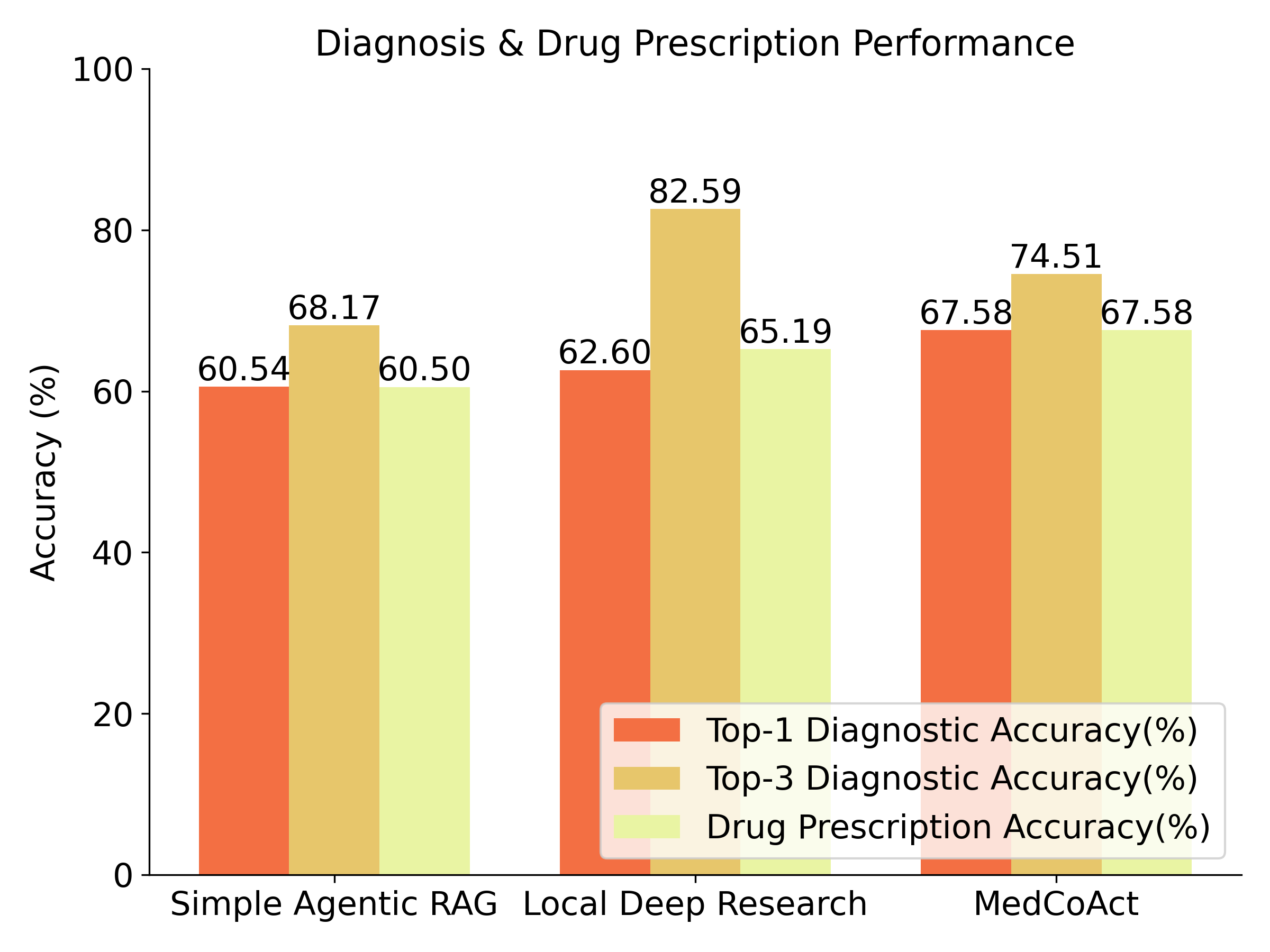}
    \caption{Accuracies of top-1 diagnostic accuracy, top-3 diagnostic accuracy, and drug prescription accuracy compared across MedCoAct and the baselines.}
    \vspace{-6mm}
    \label{fig:table2}

\end{figure}
We use \textbf{Qwen-max-0428} for experiments. Results show the effectiveness of MedCoAct on diagnostic and prescription tasks. As shown in Figure~\ref{fig:table2}, MedCoAct achieves superior performance on two of three evaluation metrics, outperforming both baseline methods in Top-1 diagnostic accuracy and drug prescription accuracy with significant improvements.
\paragraph{Limitation}
Local deep research achieves the best performance in Top-3 diagnostic accuracy at 82.59\%, compared to MedCoAct's 74.51\%. 
This superior performance can be attributed to its comprehensive web-based information gathering capability through iterative research workflows, which enables access to a broader range of medical literature and case study resources via internet search. 
However, local deep research's internet dependency limits its clinical applicability. 
Hospitals typically require closed-source knowledge bases for security and compliance, which both MedCoAct and simple agentic RAG provide. 
MedCoAct demonstrates clear clinical value within these practical constraints.
\vspace{-1mm}
\subsection{Capabilities analysis}
\paragraph{Document retrieval performance evaluation}
We utilize the LLM-as-a-judge method to calculate relevance and contribution for evaluating the quality of documents retrieved by the MedCoAct framework.
\vspace{-3mm}
\begin{table}[htbp]
\centering
\small
\caption{Performance of Medical Document Retrieval}
\label{tab:retrieval_score}
\begin{tabular}{lcc}
\toprule
\textbf{Agent} & \textbf{Relevance (0-10)} & \textbf{Contribution (0-10)} \\
\midrule
Doctor Agent & 7.14 & 5.89 \\
Pharmacist Agent & 7.45 & 6.58 \\
\bottomrule
\vspace{-3mm}
\end{tabular}
\end{table}
\vspace{-1mm}
As shown in Table~\ref{tab:retrieval_score}, both agents achieve relevance scores above 7 and contribution scores above 5, indicating effective retrieval of highly relevant and clinically valuable documents. 
Notably, the pharmacist agent shows particularly strong contribution performance, demonstrating that retrieval approaches guided by diagnostic outcomes and other medical terminology are more effective.

Both agents show higher relevance scores than contribution scores. 
This gap reveals a critical challenge: while medical retrieval systems excel at identifying relevant documents, translating this relevance into actionable clinical insights remains difficult. 
The smaller gap in the pharmacist agent indicates that specialization improves both absolute performance and the ability to extract information with diagnostic support.

\paragraph{Document retrieval performance comparison}
MedCoAct outperforms single agent baselines in retrieval quality.
To purely test document retrieval capabilities, we designed a controlled experiment using the Qwen3-4B model with relatively limited inherent medical knowledge, forcing it to perform diagnosis and drug selection strictly based on retrieved documents. 
Results show that MedCoAct improves top-1 diagnostic accuracy, top-3 diagnostic accuracy, and drug prescription accuracy over simple agentic RAG. 
These consistent performance improvements across tasks confirm that the MedCoAct collaborative retrieval mechanism can obtain higher quality medical supporting documents, effectively addressing the limitations of single agent approaches.
\begin{figure}[t]
    \centering
    \includegraphics[width=0.45\textwidth]{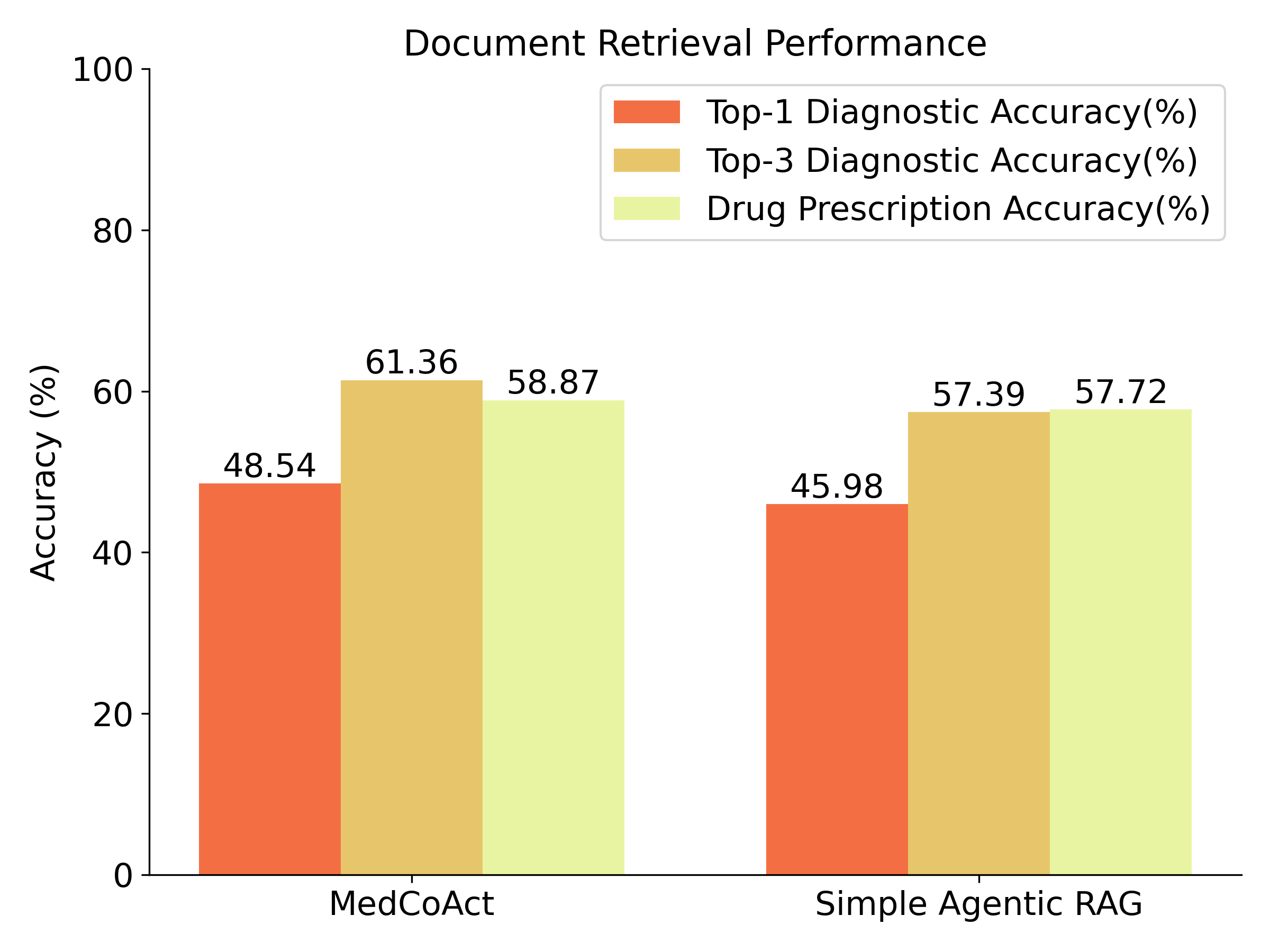}
    \caption{Accuracies of Qwen3-4B when responding to patient complaints using documents retrieved by MedCoAct and Single Agentic RAG respectively.}
    \label{fig:table4}
    \vspace{-5mm}
\end{figure}

\paragraph{Agent role specialization validation}
MedCoAct exhibits superior professional specialization and complementary role differentiation.
ROUGE-1, ROUGE-2, and ROUGE-L analysis reveals minimal overlap between doctor and pharmacist retrieved documents.
This low overlap confirms successful specialization: doctor agent targets diagnostic information while pharmacist agent focuses on medication aspects, creating complementary retrieval patterns. 
These findings validate that the dual agent architecture achieves genuine professional division of labor, demonstrating the necessity and effectiveness of role-based specialization in medical information retrieval.

\begin{figure}[htbp]
    \centering
    \vspace{0.3cm}
    \includegraphics[width=0.47\textwidth]{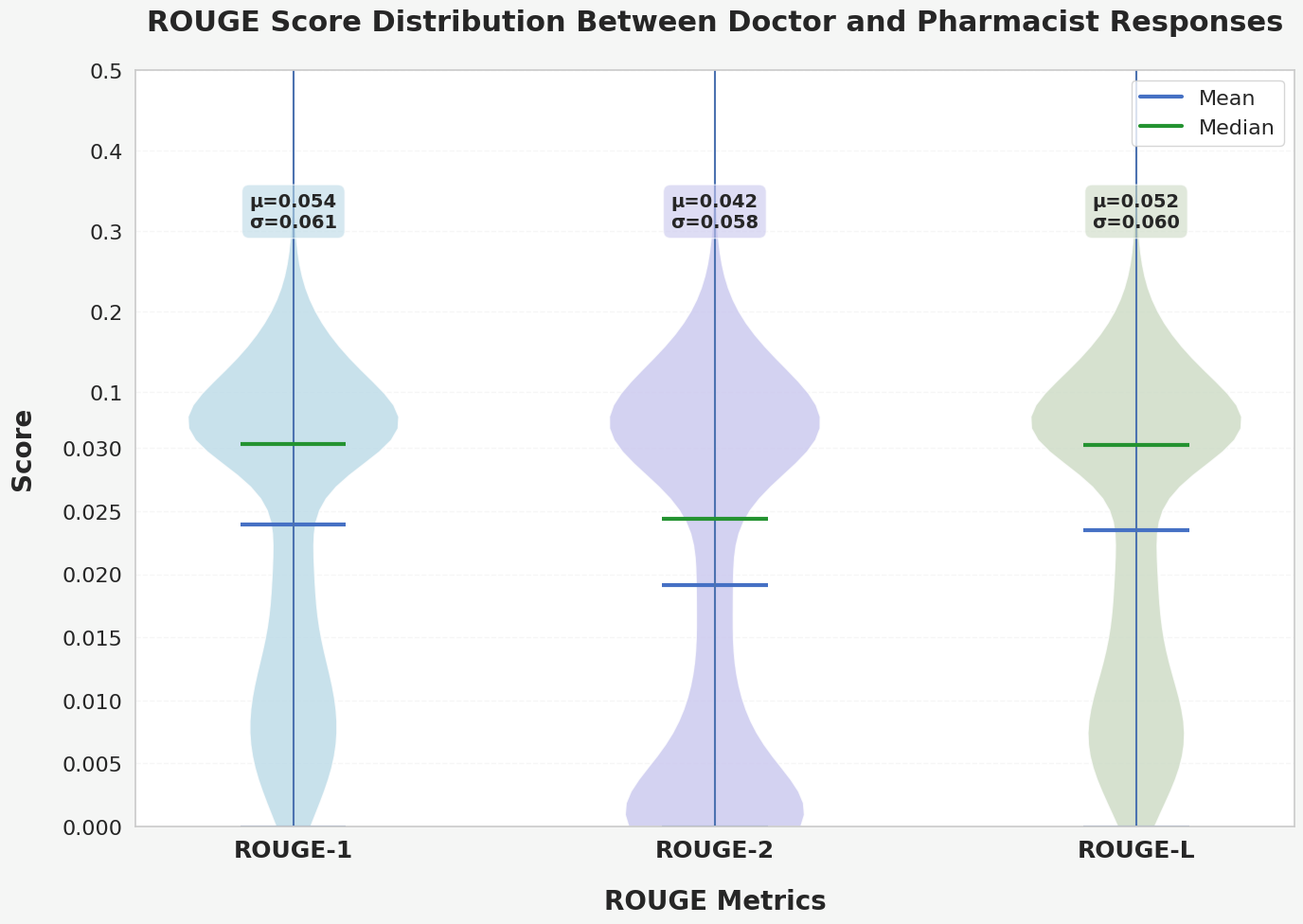}
    \caption{ROUGE score distribution between doctor and pharmacist extracted documents.}
    \label{fig:rouge_distribution}
    \vspace{-4mm}
\end{figure} 

\subsection{Failure analysis of the MedCoAct framework}
To understand why the MedCoAct framework fails, we randomly selected 50 failure cases for analysis.
Additionally, we introduced the weaker Qwen3-8B model as a comparative baseline to examine whether differences in underlying LLM capabilities would exacerbate these failure patterns. \par

Systematic analysis of failure cases reveals three primary failure modes. 
Initially, agents suffer from insufficient medical knowledge, misinterpreting key medical terms, showing inflexible responses to complex cases without adequate medication guidance and safety awareness. 
Besides, agents exhibit excessive dependence on retrieved documents, mechanically copying guidelines while ignoring patient-specific conditions, or getting trapped by retrieval results and overlooking obvious diagnostic options. 
Additionally, when facing conflicting information from multiple sources, agents lack evidence discrimination ability and simply concatenate information, producing contradictory reasoning and obviously infeasible treatment plans. 
These failure patterns worsen with the weaker Qwen3-8B model, causing more frequent reasoning collapse and risk oversight, demonstrating that framework reliability directly depends on the underlying LLM capabilities.
\vspace{-1mm}
\section{Ablation experiments}
\vspace{-1mm}
To evaluate the impact of each component of MedCoAct, we conducted ablation experiments on DrugCareQA.
\vspace{-3mm}
\begin{table}[htbp]
    \centering
    \small
    \caption{Ablation study on agents}
    \label{tab:agent_performance_final}
    \setlength{\tabcolsep}{5pt}
    \begin{tabular}{ccccc}
        \toprule
         &  & \textbf{Top-1} & \textbf{Top-3} & \textbf{Drug} \\
        \textbf{Doctor} & \textbf{Pharmacist} & \textbf{diagnostic} & \textbf{diagnostic} & \textbf{prescription} \\
        \textbf{Agent} & \textbf{Agent} & \textbf{accuracy} & \textbf{accuracy} & \textbf{accuracy} \\
         &  & \textbf{(\%)} & \textbf{(\%)} & \textbf{(\%)} \\
        \midrule
         \ding{51} &  \ding{51} & \textbf{67.58} & \textbf{74.51} & \textbf{67.58} \\
         \ding{51} &  \ding{55}& 67.58 & 74.51 & 66.43 \\
        \ding{55} &  \ding{51} & 66.02 & 73.84 & 65.91 \\
         \ding{55} &  \ding{55} & 65.91 & 73.88 & 63.69 \\
        \bottomrule
    \end{tabular}
    \vspace{-3mm}
\end{table}
\vspace{-2mm}
\subsection{The effectiveness of agents}
\vspace{-1mm}
To understand the impact of different agents on the final results, we exclude certain agents and replace them with naive RAG. As indicated by Table~\ref{tab:agent_performance_final}, the addition of specialized agents different from just naive RAG consistently improves both diagnostic and prescription accuracy. Notably, drug prescription tasks are more sensitive to component removal than diagnostic tasks, which we attribute to the diagnosis-first-then-prescription workflow design where prescription accuracy depends heavily on preceding diagnostic results. While specialized agents slightly increase computational complexity, the overall performance improves noticeably, demonstrating the effectiveness of the agent-based approach.
\vspace{-2mm}
\subsection{The effectiveness of collaborative agent mechanism}
As shown in Figure~\ref{fig:table4}, implementing collaboration of doctor agent and pharmacist agent in MedCoAct leads to significant improvements over single agent method across all metrics. 
Besides, Table~\ref{tab:agent_performance_final} shows that even the naive dual RAG architecture enhances the performance compared to single RAG method. 
These results demonstrate that our designed multi-agent collaboration mechanism can significantly improve medical decision accuracy through specialized division of labor, effectively reducing information loss and bias, thereby improving overall system performance.
\vspace{-1mm}
\section{Conclusion}
We identify a critical limitation where current medical AI systems process diagnostic and medication tasks in isolation, lacking collaborative mechanisms and resulting in suboptimal real-world performance. To address this, we introduce MedCoAct, a confidence-aware multi-agent framework that simulates doctor-pharmacist collaboration in integrated diagnosis-to-treatment workflows. To evaluate our approach, we introduce DrugCareQA, an evaluation dataset specifically designed for integrated clinical workflows. Experimental results show that our collaborative framework significantly outperforms baseline methods on both diagnostic and medication tasks through confidence-aware reflection mechanisms and interpretable decision-making pathways via role specialization. Future work can extend MedCoAct to broader specialties and explore advanced inter-agent communication for healthcare integration.
\vspace{-1mm}
\bibliographystyle{IEEEtran}
\bibliography{IEEEexample}
\end{document}